\documentclass[11pt,a4paper]{article}
\usepackage[hyperref]{emnlp2020}
\usepackage{pslatex}
\usepackage{latexsym}
\usepackage{amsmath} 
\usepackage{CJKutf8}
\usepackage{booktabs}
\usepackage{makecell}
\usepackage{graphicx}
\usepackage{multirow}
\usepackage{subfig}
\usepackage{url}
\usepackage{bibentry}
\usepackage{amssymb}
\usepackage{cleveref}
\usepackage[utf8]{inputenc}

\definecolor{ao}{rgb}{0.0, 0.5, 0.0}

\usepackage{microtype}

\aclfinalcopy % Uncomment this line for the final submission

\title{Bootstrapping a Crosslingual Semantic Parser}

\author{Tom Sherborne, Yumo Xu {\rm and } Mirella Lapata\\
  Institute for Language, Cognition and Computation \\
  School of Informatics, University of Edinburgh \\
  10 Crichton Street, Edinburgh EH8 9AB \\
  \texttt{\{tom.sherborne,yumo.xu\}@ed.ac.uk, mlap@inf.ed.ac.uk} \\
}

\date{}

\begin{document}
\setlength{\abovedisplayskip}{2pt}
\setlength{\belowdisplayskip}{2pt}

\maketitle
\begin{abstract}

  Recent progress in semantic parsing scarcely considers languages
  other than English but professional translation can be prohibitively
  expensive. We adapt a semantic parser trained on a single language,
  such as English, to new languages and multiple domains with minimal
  annotation. We query if machine translation is an adequate
  substitute for training data, and extend this to investigate
  bootstrapping using joint training with English, paraphrasing, and
  multilingual pre-trained models. We develop a Transformer-based
  parser combining paraphrases by ensembling attention over multiple
  encoders and present new versions of ATIS and Overnight in German
  and Chinese for evaluation.  Experimental results indicate that MT
  can approximate training data in a new language for accurate parsing
  when augmented with paraphrasing through multiple MT
  engines. Considering when MT is inadequate, we also find that using
  our approach achieves parsing accuracy within 2\% of complete
  translation using only 50\% of training data.\footnote{Our code and
    data can be found at \url{github.com/tomsherborne/bootstrap}.  }

\end{abstract}

\section{Introduction}
\label{sec:introduction}

Semantic parsing is the task of mapping natural language utterances to
machine-interpretable expressions such as SQL or a logical meaning
representation. This has emerged as a key technology for developing
natural language interfaces, especially in the context of question
answering
\cite{kwiatkowski-etal-2013-scaling,berant-etal-2013-semantic,Liang:2016,kollar-etal-2018-alexa},
where a semantically complex question is translated to an executable
{\em query} to retrieve an answer, or {\em denotation}, from a
knowledge base.

%We can use semantic parsing to
%facilitate a natural language interface to databases (NLIDB) wherein a
%semantic parser translates from natural language to an executable
%logical form, or {\em query}, proceeds to execute such logical form to
%retrieve an answer, or {\em denotation}.

Sequence-to-sequence neural networks
\citep{seq2seq-DBLP:journals/corr/SutskeverVL14} are a popular
approach to semantic parsing, framing the task as {\em sequence
  transduction} from natural to formal languages
\citep{attn-copy-Jia2016, seq2tree-Dong2016-lang2logic}. Recent
proposals include learning intermediate logic representations
\citep{coarse2fine-Dong2018, Guo2019TowardsCT-IRnet}, constrained
decoding \citep{yin-neubig-2017-syntactic-gcd-code-geneeration,
  krishnamurthy-etal-2017-gcd-entity-link-wikitabqs,
  lin2019-grammar-decoding-text-to-sql}, and graph-based parsing
\citep{Bogin2019RepresentingSS-gnn-text-to-sql-acl19,
  Shaw2019GeneratingLF-gnn-semparse-acl19}.

Given recent interest in semantic parsing and the
data requirements of neural methods, it is unsurprising that
many challenging datasets have been released in the past decade
\citep{Overnight-Wang15, zhongSeq2SQL2017-wikisql,
  iyer-etal-2017-learning-user-feedback, spider-Yu2018,
  Yu2019SParCCS-semparse-across-conversation}. However, these
widely use English as synonymous for natural language. English is
neither linguistically typical \citep{wals} nor the most widely
spoken language worldwide \citep{ethnologue}, but is presently the
{\em lingua franca} of both utterances and knowledge bases in semantic
parsing. Natural language interfaces intended for international
deployment must be adaptable to multiple locales beyond prototypes for
English. However, it is uneconomical to create brand new 
datasets for every new language and domain.

In this regard, most previous work has focused on \emph{multilingual}
semantic parsing i.e.,~learning from multiple natural languages in
parallel assuming the availability of multilingual training data.
Examples of multilingual datasets include GeoQuery
\citep{geoquery-Zelle:1996:LPD:1864519.1864543}, ATIS
\citep{atis-Dahl:1994:ESA:1075812.1075823} and NLMaps
\citep{nlmaps-Haas2016} but each is limited to one domain. For larger
datasets, professional translation can be prohibitively expensive and
require many man-hours from experts and native speakers.  Recently,
\citet{Min2019-CSPIDER} reproduced the public partitions of the SPIDER
dataset \citep{spider-Yu2018} into Chinese, but this required three
expert annotators for verification and agreement.  We posit  there
exists a more efficient strategy for expanding semantic parsing to a
new language.

% There presently exists a scarcity of datasets for multi-domain
% executable semantic parsing from a language other than English. Most
% work in multilingual semantic parsing has studied GeoQuery
% \citep{geoquery-Zelle:1996:LPD:1864519.1864543}, ATIS
% \citep{atis-Dahl:1994:ESA:1075812.1075823} and NLMaps
% \citep{nlmaps-Haas2016}, which each cover only a single domain of
% information.

In this work, we consider {\em crosslingual semantic parsing},
adapting a semantic parser trained on English, to another language.
We expand executable semantic parsing to new languages and multiple
domains by bootstrapping from in-task English datasets, task-agnostic
multilingual resources, and publicly available machine translation
(MT) services, in lieu of expert translation of training data. We
investigate a core hypothesis that MT can provide a noisy, but
reasonable, approximation of training data in a new source
language. We further explore the benefit of augmenting noisy MT data
using pre-trained models, such as BERT \citep{devlin-etal-2019-BERT},
and multilingual training with English. Additionally, we examine
approaches to ensembling multiple machine translations as approximate
paraphrases. This challenge combines both \textit{domain adaptation}
and \textit{localization}, as a parser must generalize to the
locale-specific style of queries using only noisy examples to learn
from.

%In this work, we propose a recourse to expand multi-domain, executable
%semantic parsing to new languages by bootstrapping using existing
%resources and {\em minimally hands-on}, partial dataset translation
%without domain expertise. The benefits of our approach are
%twofold. Firstly, this economises on existing resources and enables
%further crosslingual language understanding (XLU) for semantic
%parsing, similar to the approach of \citet{conneau2018xnli} for
%entailment. Secondly, we propose a cost-effective translation
%methodology to enable language expansion of benchmark datasets and QA
%systems used in commercial systems.

For our evaluation, we present the first \emph{multi-domain},
executable semantic parsing dataset in three languages and an
additional locale for a \emph{single-domain} dataset.  Specifically,
we extend ATIS \citep{atis-Dahl:1994:ESA:1075812.1075823}, pairing
Chinese (ZH) utterances from
\citet{arch-for-neural-multisp-Susanto2017} to SQL queries and create
a parallel German (DE) human-translation of the full dataset.
Following this, we also make available a new version of the
multi-domain Overnight dataset \citep{Overnight-Wang15} where only
development and test sets are translations from native speakers of
Chinese and German. This is representative of the real-world scenario
where a semantic parser needs to be developed for new languages
without gold-standard training data.

Our contributions can be summarized as follows: (1)~new versions of
ATIS \citep{atis-Dahl:1994:ESA:1075812.1075823} and Overnight
\citep{Overnight-Wang15} for generating executable logical forms from
Chinese and German utterances; (2)~a combined encoder-decoder
attention mechanism to ensemble over multiple Transformer encoders;
(3)~a cost-effective methodology for bootstrapping semantic parsers to
new languages using minimal new annotation. Our proposed method
overcomes the paucity of gold-standard training data using pre-trained
models, joint training with English, and paraphrasing through MT
engines; and~(4)~an investigation into practical minimum gold-standard
translation requirements for a fixed performance penalty when MT is
unavailable.

\begin{table*}[t!]
\centering
\begin{tabular}{@{}ll@{}}
\toprule
\multicolumn{2}{l}{Noun/Adjective Ambiguity (``first-class fares'' is a noun object)} \\ \midrule
EN & Show me the first class fares from Baltimore to Dallas \\
DE$_{\text{MT}}$ & Zeigen Sie mir die {\color{red} erstklassigen} Tarife von Baltimore nach Dallas \\
DE$_{\text{H}}$ & Zeige mir die Preise in der {\color{ao} ersten
                  Klasse} von Baltimore nach Dallas \\\midrule
\multicolumn{2}{l}{Entity Misinterpretation (Airline names aren't preserved)} \\ \midrule
EN & Which Northwest and United flights go through Denver before noon? \\ 
DE$_{\text{MT}}$ & Welche {\color{red} Nordwesten} und {\color{red} Vereinigten }Flüge gehen durch Denver vor Mittag \\
DE$_{\text{H}}$ & Welche {\color{ao}Northwest} und {\color{ao}United }Flüge gehen durch Denver vor Mittag \\ \midrule
\multicolumn{2}{l}{Question to Statement Mistranslation (rephrased as ``You have a...'')} \\ \midrule
EN & Do you have an 819 flight from Denver to San Francisco? \\
ZH$_{\text{MT}}$ & \begin{CJK*}{UTF8}{gbsn}{\color{red}你有}一个从丹佛到旧金山的 819 航班 \end{CJK*} \\
ZH$_{\text{H}}$ & \begin{CJK*}{UTF8}{gbsn}{\color{ao}有没有}从丹佛到旧金山的 819 航班\end{CJK*} \\ \midrule
\multicolumn{2}{l}{Contextual Misinterpretation (``blocks'' translated to \begin{CJK*}{UTF8}{gbsn}``街区''\end{CJK*} [street blocks)])} \\ \midrule
EN & What seasons did Kobe Bryant have only three blocks? \\
ZH$_{\text{MT}}$ & \begin{CJK*}{UTF8}{gbsn}什么季节科比布莱恩特只有三个{\color{red}街区}\end{CJK*} \\ \midrule
\multicolumn{2}{l}{Referential Ambiguity (\begin{CJK*}{UTF8}{gbsn}他\end{CJK*}[he] refers to either players or Kobe Bryant)} \\ \midrule
EN & Which players played more games than Kobe Bryant the seasons he played? \\
ZH$_{\text{MT}}$ & \begin{CJK*}{UTF8}{gbsn}在{\color{red}他}打球的那
些赛季中,哪些球员比科比布莱恩特打得更多\end{CJK*} \\ 
\bottomrule
\end{tabular}
\caption{Examples from ATIS \protect \citep{atis-Dahl:1994:ESA:1075812.1075823}
  and Overnight  \protect \cite{Overnight-Wang15}. Utterances are translated 
  into Chinese and German using both machine translation (L$_{\text{MT}}$) and
  crowdsourcing with verification (L$_{\text{H}}$). We highlight some issues with the noisy MT data ({\color{red}red}), contrasting to improved human translations ({\color{ao}green}) for ATIS.} 
\label{tab:dset_examples}
\end{table*}

%%%%%%%%%%%%%%%%%%%%%%%%%%%%%%%%%%%%%%%%%%%%%%%%%%%%%%%%%%%%%%%%%%%%%5
\section{Related Work}
\label{sec:related-work}

Across logical formalisms, there have been several proposals for
multilingual semantic parsing which employ multiple natural languages
in parallel
\citep{geoquery-german-bevan-Jones:2012:SPB:2390524.2390593,
  andreas-etal-2013-semantic, lu-2014-semantic-hybrid-trees,
  neural-hybrid-trees-susanto2017,
  jie-lu-2018-dependency-hybrid-trees-semparse}.

\citet{multilingual-sp-hierch-tree-Jie2014} ensemble monolingual
parsers to generate a single parse from $<5$ source languages for
GeoQuery \cite{geoquery-Zelle:1996:LPD:1864519.1864543}. Similarly,
\citet{polyglotapi-Richardson2018} propose a polyglot automaton
decoder for source-code generation in 45
languages. \citet{arch-for-neural-multisp-Susanto2017} explore a
multilingual neural architecture in four languages for GeoQuery and
three languages for ATIS by extending
\citet{seq2tree-Dong2016-lang2logic} with multilingual encoders.
Other work focuses on multilingual representations for semantic
parsing based on universal dependencies \cite{reddy-etal-2017-universal-semparse} or embeddings of logical forms \cite{learning-xling-reps-for-sp-Zou2018}.

We capitalize on existing semantic parsing datasets to bootstrap from
English to another language, and therefore, do not assume that
multiple languages are available as parallel input. Our work is
closest to \citet{multilingsp-and-code-switching-Duong2017}, however
they explore how to parse both English and German simultaneously using
a multilingual corpus. In contrast, we consider English data only as
an augmentation to improve parsing in Chinese and German and do not
use ``real'' utterances during training. Recently,
\citet{artetxe2020translation-artifacts} studied MT for crosslingual
entailment, however, our results in Section \ref{sec:results} suggest
these prior findings may not extend to semantic parsing, owing to the
heightened requirement for factual consistency across
translations. 

Our work complements recent efforts in crosslingual language
understanding such as XNLI for entailment \cite{conneau2018xnli},
semantic textual similarity \cite{cer-etal-2017-semeval} or the XTREME
\cite{hu2020xtreme} and XGLUE \citep{liang2020xglue} benchmarks. There
has also been interest in parsing into interlingual graphical meaning
representations \citep{Damonte2018-xlingual-amr,
  xling-sp-acl-Zhang2018a}, spoken language understanding
\citep{8461905} and $\lambda$-calculus expressions
\citep{Kwiatkowski:2010:IPC:1870658.1870777-semparsehigherorder,
  geoquery-zh-lu-ng-2011-probabilistic,
  lu-2014-semantic-hybrid-trees}.  In contrast, we focus on logical
forms grounded in knowledge-bases and therefore do not consider these
approaches further.

\section{Problem Formulation}
\label{sec:problem-formulation}

Throughout this work, we consider the real-world scenario where a
typical developer wishes to develop a semantic parser to
facilitate question answering from an existing commercial database to customers in a
new locale. For example, an engineer desiring to extend support to German speakers for a commercial database of USA flights in English. Without the resources of high-valued technology companies, costs for annotation and machine learning resources must be
minimized to maintain commercial viability. To economize this task, the developer must minimize new annotation or professional translation and instead bootstrap a system with public resources. At a minimum, a test and development set of
utterances from native speakers are required for evaluation. However,
the extent of annotation and the utility of domain adaptation for training are
unknown. Therefore, our main question is {\it how successfully can a
semantic parser learn with alternative data resources to generalize
to novel queries in a new language?}

Crosslingual semantic parsing presents a unique challenge as an NLU
task. It demands the generation of precise utterance semantics,
aligned across languages while ensuring an accurate mapping between
logical form and the idiomatic syntax of questions in every language
under test.  In comparison to NLU classification tasks such as XNLI
\citep{conneau2018xnli}, our challenge is to \textbf{preserve} and
\textbf{generate} meaning, constrained under a noisy MT channel.  The
misinterpretation of entities, relationships, and relative or
numerical expressions can all result in an incorrect parse.  

Lexical translation in MT, however accurate it may be, is insufficient
alone to represent queries from native speakers. For example, the
English expression ``dinner flights'' can be directly translated to
German as ``Abendessenflug'' [dinner flight], but ``Flug zur
Abendszeit'' [evening flight] better represents typical German
dialogue. This issue further concerns question phrasing. For example,
the English query ``do you have X?'' is often mistranslated to a
statement
\begin{CJK*}{UTF8}{gbsn}``你有一个X''\end{CJK*} [you have one X] but
typical Chinese employs a positive-negative pattern
(\begin{CJK*}{UTF8}{gbsn}``有没有一个X?''\end{CJK*} [have not have one
X?])  to query possession. Our parser must overcome each of these
challenges without access to gold data.

\subsection{Neural Semantic Parsing}
\label{ssec:neural_sp}

We approach our semantic parsing task using a {\sc Seq2Seq}
architecture \textit{Transformer} encoder-decoder
network \citep{transformers-noam-Vaswani2017AttentionIA}. The encoder
computes a contextual representation for each input token through
\textit{multi-head self-attention} by combining parallel dot-product
attention weightings, or ``heads'', over the input sequence.  The
decoder repeats this self-attention across the output sequence and
incorporates the source sequence through multi-head attention over the
encoder output. A Transformer layer maps input $X =
\{x_{i}\}_{i=0}^{N}$, where $x_{i}\in \mathbb{R}^{d_{x}}$, to output
\mbox{$Y = \{y_{i}\}_{i=0}^{N}$} using attention components of
Query~$\bf Q$, Key $\bf K$ and Value $\bf V$ in $H$ attention heads:

\begingroup\makeatletter\def\f@size{10}\check@mathfonts
\def\maketag@@@#1{\hbox{\m@th\large\normalfont#1}}%
\begin{align} 
  {\bf e}_{i}^{(h)} =& \frac{{\bf Q}W_{Q}^{(h)}\left({\bf
        K}W_{K}^{(h)}\right)^{T}}{\sqrt{d_{x}/H}};~{\bf s}_{i}^{(h)}=\operatorname{softmax}\left({\bf e}_{i}^{(h)}\right)\hspace*{-.4cm} \label{eq:selfattn} \\[5pt]
    {\bf z}_{i}^{(h)} =& {\bf s}_{i}^{(h)}\left({\bf V}W_{V}^{(h)}\right); ~~{\bf z}_{i}=\operatorname{concat}\{{\bf
      z}_{i}^{(h)}\}_{h=1}^{H} \label{eq:transformer}  \\[5pt] 
   \hat{{\bf y}}_{i}=&\operatorname{LayerNorm}\left(X+{\bf
        z}_{i}\right) \label{eq:yhat} \\[5pt] 
   \mathbf{y}_{i}=&\operatorname{LayerNorm}\left(\hat{{\bf y}}_{i} + {\rm
        FC}\left( \operatorname{ReLU} \left(  {\rm FC } \left(\hat{{\bf y}}_{i}
          \right) \right) \right) \right) \label{eq:yout}  
\end{align}\endgroup
%\vspace{.25em}

Following \citet{wang2019ratsql}, Equation~\ref{eq:selfattn} describes
attention scores between Query ($Q$) and Key ($K$), ${\bf z}_{i}^{h}$
is the $h^{\rm th}$ attention head, applying scores ${\bf
  s}_{i}^{(h)}$ to value ($V$) into the multi-head attention function
${\bf z}_{i}$ with $W^{(h)}_{\{Q,K,V\}} \in
\mathbb{R}^{d_{x}\times(d_{x}/H)}$. Output prediction ${\bf y}_{i}$ combines
${\bf z}_{i}$ with a residual connection and two fully-connected (FC)
layers, ReLU nonlinearity, and layer normalization
\citep{ba2016layer}. The encoder computes self-attention through
query, key, and value all equal to the input, $\{{\bf Q}, {\bf K},
{\bf V}\}=X$.  Decoder layers use self-attention over output sequence,
$\{{\bf Q}, {\bf K}, {\bf V}\}=Y_{out}$, followed by attention over
the encoder output~$E$ (${\bf Q}=Y_{out}$ and $\{{\bf K},{\bf
  V}\}=E$) to incorporate the input encoding into decoding.

\begin{figure}[t]
    \centering
    \includegraphics[width=0.4\textwidth]{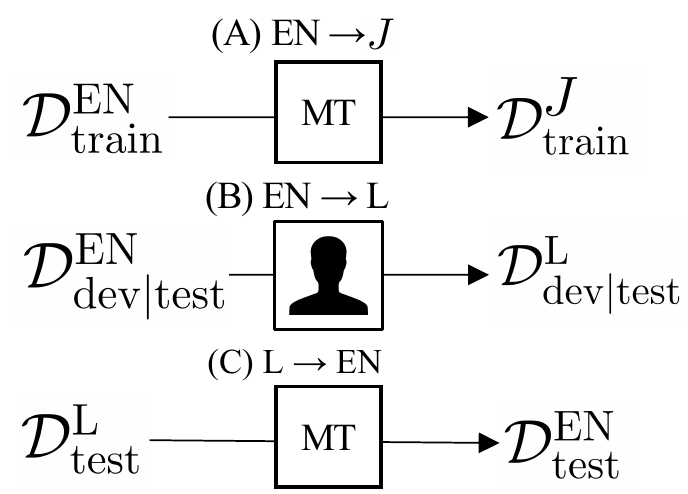}
    \caption{(A) Machine Translation (MT) from English into some
      language, L, for training data. $J$~is the MT approximation of
      this language to be parsed. (B) Human translation of the
      development and test sets from English into language L. (C)
      Translation from language L~into English using MT. Any system
      parsing language L~must perform above this ``back-translation''
      baseline to justify development.} 
    \label{fig:data_generation_examples}
\end{figure}

\subsection{Crosslingual Modeling}
\label{sec2:xlingual_modelling}

Consider a parser, ${\rm SP}\left(x\right)$, which transforms
utterances in language~$x_{\rm L}$, to some executable logical form,
$y$. We express a dataset in some language~$\rm L$ as
$\mathcal{D}^{~\rm L} = \left(\{x^{~\rm 
L}_{n},~y_{n},~d_{n}\}^{N}_{n=1}, KB\right)$, for $N$
examples where $x^{~\rm L}$ is an utterance in language ${\rm L}$, $y$ is the corresponding logical form and $d$ is a denotation from knowledge base,
$d=KB\left(y\right)$. The MT approximation of language $\rm
L$ is described as $J$; using MT from English, $x^{J}={\rm
MT}\left(x^{\rm EN} \right)$. Our hypothesis is that $J \approx {\rm L}$ such that prediction $\hat{y} = {\rm SP}\left(x^{\rm L}\right)$ for test example $x^{\rm L}$ approaches gold logical form, $y_{\rm gold}$, conditioned
upon the quality of MT. An ideal parser will output non-spurious prediction, $\hat{y}$, executing to return an equal denotation to $KB\left(y_{\rm gold}\right)=d_{\rm gold}$. The proportion of predicted queries which retrieve the correct denotation defines the \textit{denotation accuracy}. Generalization performance is always measured on real queries from native speakers e.g.~
$\mathcal{D}^J=\lbrace \mathcal{D}^{J}_{\rm train}, \mathcal{D}^{\rm L}_{\rm dev} , \mathcal{D}^{\rm L}_{\rm test} \rbrace$ and
$\mathcal{D}^{J}_{\rm dev|test} = \emptyset$.

We evaluate parsing on two languages to compare transfer
learning from English into varied locales. We investigate German,
a similar Germanic language, and Mandarin Chinese, a dissimilar
Sino-Tibetan language, due to the purported quality of existing MT
systems \cite{gtranslate} and availability of native speakers to
verify or rewrite crowdsourced annotation. 
Similar to \citet{conneau2018xnli}, we implement a ``back-translate into English'' baseline wherein the test set in ZH/DE is machine translated into English and a semantic parser trained on the source English dataset predicts logical forms. Figure~\ref{fig:data_generation_examples} indicates how each dataset is generated. To maintain a commercial motivation for developing an in-language parser, any proposed system must perform above this baseline. Note that we do not claim to be investigating
semantic parsing for low-resource languages since, by virtue, we
require adequate MT into each language of interest. We use Google Translate \cite{gtranslate} as our primary MT
system and complement this with systems from other global
providers. The selection and use of MT is further discussed in
Appendix~\ref{sec:appendix_data}.

\begin{figure}[t]
    \centering
    \includegraphics[width=0.4\textwidth]{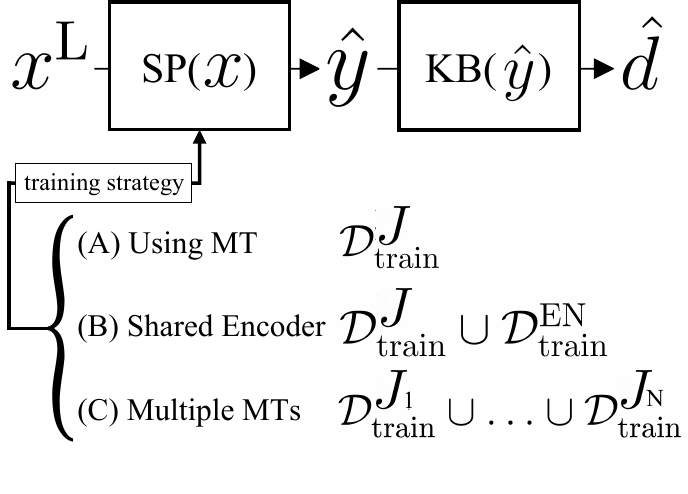}
%\vspace*{-2ex}
    \caption{The semantic parser (SP) predicts a logical form,
      $\hat{y}$, from an utterance in language L, $x^{\rm L}$. A
      knowledge base (KB) executes the logical form to predict a
      denotation, $\hat{d}$. Approaches to crosslingual modeling
      involve: (A)~using machine translation (MT) to approximate
      training data in language L; (B)~training SP on both MT data and
      source English data; (C)~using multiple MT systems to improve
      the approximation of~L.}
    \label{fig:xlingual_modelling}
\end{figure}

\subsection{Feature Augmentation}
\label{sec:feat_aug}
Beyond using MT for in-language training data, we now describe our
approach to further improve parsing using external resources and
transfer learning. These approaches are described in Figure~\ref{fig:xlingual_modelling}.

\paragraph{Pre-trained Representations}

Motivated by the success of contextual word representations for
semantic parsing of English by
\citet{Shaw2019GeneratingLF-gnn-semparse-acl19}, we extend this
technique to Chinese and German using implementations of BERT from
\citet{Wolf2019HuggingFacesTS}. Rather than learning embeddings for
the source language {\em tabula rasa}, we experiment with using
pretrained 768-dimensional inputs from BERT-base in English, Chinese and German\footnote{{\tt deepset.ai/german-bert}}, as well as the multilingual model trained on 104 languages. To account for rare entities which may be absent from pre-trained vocabularies, we append these representations to learnable embeddings. Representations for logical form tokens 
are trained from a random initialisation, as we lack a BERT-style pre-trained model for meaning representations (i.e.,~$\lambda-$DCS or SQL queries). Early experiments considering multilingual word representations
\cite{conneau2017word-MUSE,song-etal-2018-directional-zh-embeds} yielded no significant improvement and these results are omitted for brevity.

\paragraph{Multilingual ``Shared'' Encoder}

Following \citet{multilingsp-and-code-switching-Duong2017} and
\citet{arch-for-neural-multisp-Susanto2017}, we experiment with an
encoder trained with batches from multiple languages as input. 
Errors in the MT data are purportedly mitigated through the
model observing an equivalent English utterance for the same logical
form. The joint training dataset is described as
$\mathcal{D}_{\rm train}^{{\rm EN}+{J}}=\mathcal{D}_{\rm train}^{\rm
  EN}\cup \mathcal{D}_{\rm train}^{{J}}$
for $J=\{{\rm ZH},{\rm DE}\}$. Consistent with Section \ref{sec2:xlingual_modelling}, we measure validation and test performance using only utterances from native speakers,
$\mathcal{D}_{\rm dev|test}^{\rm L}$, and ignore performance for
English. This is similar to the {\tt All} model from
\citet{multilingsp-and-code-switching-Duong2017}, however, our objective
is biased to maximize performance on one language rather
than a balanced multilingual objective.

\paragraph{Machine Translation as Paraphrasing}
Paraphrasing is a common augmentation for semantic parsers to improve
generalization to unseen utterances
\citep{sp-via-paraphrasing-Berant2014,
  dong-etal-2017-learning-to-paraphrase,
  iyer-etal-2017-learning-user-feedback,
  su-yan-2017-xdomain-paraphrasing, 2018arXiv180400401U}. While there has been some study
of multilingual paraphrase systems
\citep{ganitkevitch-callison-burch-2014-multilingual-ppdb}, we instead
use MT as a paraphrase resource, similar to
\citet{mallinson-etal-2017-paraphrasing}. Each MT system will have
have different outputs from different language models and therefore we
hypothesize that an ensemble of multiple systems,
$\left(J_{1}, \ldots J_{N}\right)$, will provide greater linguistic
diversity to better approximate~${\rm L}$.  Whereas prior work uses
back-translation or beam search, a developer in our scenario lacks the
resources to train a NMT system for such techniques. As a
shortcut, we input the same English sentence into $m$~public APIs for
MT to retrieve a set of candidate paraphrases in the language of
interest (we use three APIs in experiments).

We experiment with two approaches to utilising these
pseudo-paraphrases. The first, \mbox{MT-Paraphrase}, aims to learn a
single, robust language model for ${\rm L}$ by uniformly sampling one
paraphrase from $\left(J_{1}, \ldots J_{N}\right)$ as input to the
model during each epoch of training. The second approach,
\mbox{MT-Ensemble}, is an ensemble architecture similar to
\citet{Garmash2016-ensemble} and \citet{Firat2016-zeroshotmt}
combining attention over each paraphrase in a single decoder.  For~$N$
paraphrases, we train ~$N$ parallel encoder models,
$\{e_n\}_{n=1}^{N}$, and ensemble across each paraphrase by combining
$N$ sets of encoder-decoder attention heads.  For each encoder output,
$E_n=e_n\left(X_n\right)$, we compute multi-head attention, ${\bf
  z}_{i}$ in Equation~\ref{eq:transformer}, with the decoder state,
$D$, as the query and $E_{n}$ as the key and value
(Equation~\ref{eq:m_mha}). Attention heads are combined through a
combination function (Equation \ref{eq:m_comb}) and output~${\bf m}_{i
  \epsilon}$ replaces~${\bf z}_{i}$ in Equation \ref{eq:yhat}.  

We compare ensemble strategies using two combination functions: the
mean of heads (Equation~\ref{eq:comb}a) and a gating network
(\citealt{Garmash2016-ensemble}; Equation~\ref{eq:comb}b) with gating
function~$\mathbf{g}$ (Equation~\ref{eq:g_gate_fn}) where
$W_\mathbf{g} \in R^{N\times|V|}, W_h \in R^{|V| \times N|V|}$.  We
experimentally found the gating approach to be superior and we report
results using only this method.

\begin{align}
{\bf m}_{n} &= {\rm MultiHeadAttention}\left(D, E_{n}, E_{n}\right) \label{eq:m_mha}\\
{\bf m}_{i \epsilon} &= {\rm comb}\left({\bf m}_{1}, \ldots {\bf m}_{N}\right) \label{eq:m_comb} 
\end{align} 

\begin{equation}
\label{eq:comb}
\text{comb}=
\begin{cases}
    \frac{1}{N}\sum_{n}^{N} {\bf m}_{n} & \hfill \text{(a)}\\
\sum_{n}^{N} \mathbf{g}_{n} {\bf m}_{n} &\hfill \text{(b)} \\
\end{cases}
\end{equation}

\begin{equation}
{\bf g}  = {\rm softmax}\left( W_{\mathbf{g}} {\rm tanh}\left( W_{h} [ {\bf m}_{n}, \ldots {\bf m}_{N}] \right) \right)  \label{eq:g_gate_fn}
\end{equation}
\vspace{.1em}

Each expert submodel uses a shared embedding space to exploit
similarity between paraphrases. During training, each encoder learns a
language model specific to an individual MT source, yielding diversity
among experts in the final system. However, in order to improve
robustness of each encoder to translation variability, inputs to
each encoder are shuffled by some tuned probability
$p_{\rm shuffle}$. During prediction, the test utterance is input to
all $N$ models in parallel. In initial experiments, we found negligible
difference in \mbox{MT-Paraphrase} using random sampling or round-robin selection 
of each paraphrase. Therefore, we assume that both methods use all
available paraphrases over training. Our two approaches differ 
in that \mbox{MT-Paraphrase} uses all paraphrases sequentially 
whereas \mbox{MT-Ensemble} uses paraphrases in parallel. 
Previous LSTM-based ensemble approaches propose training full 
parallel networks and ensemble at the final decoding step. 
However, we found this was too expensive given the
non-recurrent Transformer model. Our hybrid mechanism permits the
decoder to attend to every paraphrased input and maintains a tractable
model size with a single decoder.

%%% Local Variables: 
%%% mode: latex
%%% TeX-master: "../../EMNLP2020-Bootstrap/main"
%%% End: 

\section{Data}
\label{sec:data}

We consider two datasets in this work. Firstly, we evaluate our
hypothesis that MT is an adequate proxy for ``real'' utterances using
ATIS \citep{atis-Dahl:1994:ESA:1075812.1075823}. This
\emph{single-domain} dataset contains 5,418 utterances paired with SQL
queries pertaining to a US flights database. ATIS was previously
translated into Chinese by \citet{arch-for-neural-multisp-Susanto2017}
for semantic parsing into $\lambda$-calculus, whereas we present these
Chinese utterances aligned with SQL queries from
\citet{iyer-etal-2017-learning-user-feedback}. In addition, we
translate ATIS into German following the methodology described below.
We use the split of 4,473/497/448 examples for train/validation/test
from \citet{lexicalgeneralisation-with-ccg-Kwiatkowski2011}.

We also examine the \emph{multi-domain} Overnight dataset
\citep{Overnight-Wang15}, which contains 13,682 English questions
paired with $\lambda-$DCS logical forms executable in SEMPRE
\citep{berant-etal-2013-semantic}. Overnight is 2.5$\times$ larger
than ATIS, so a complete translation of this dataset would be
uneconomical for our case study. As a compromise, we collect human
translations in German and Chinese only for the test and validation
partitions of Overnight. We argue that having access to limited
translation data better represents the crosslingual transfer
required in localizing a parser. We define a fixed development
partition of a stratified 20\% of the training set for a final split
of 8,754/2,188/2,740 for training/validation/testing. Note we consider
only Simplified Mandarin Chinese for both datasets.

\paragraph{Crowdsourcing Translations}

The ATIS and Overnight datasets were translated to German and Chinese
using Amazon Mechanical Turk, following best practices in related work
\citep{fastcheapcreative-Callison-Burch2009,
  crowdsourcingfromnonprofessionals-Zaidan2011,
  improvingeduMTwithcrowdsourcing-Behnke2018,
  multilingualeducorpus-Sosoni2018}.

We initially collected three translations per source
sentence. Submissions were restricted to Turkers from Germany,
Austria, and Switzerland for German and China, USA, or Singapore for
Chinese. Our AMT interface barred empty submissions and copying or
pasting anywhere within the page.  Any attempts to bypass these
controls triggered a warning message that using MT is prohibited.
Submissions were rejected if they were $>80$\% similar (by BLEU) to
references from Google Translate \citep{gtranslate}, as were
nonsensical or irrelevant submissions.

In a second stage, workers cross-checked translations by rating the
best translation from each candidate set, including an MT reference,
with a rewrite option if no candidate was satisfactory. We collected
three judgements per set to extract the best candidate
translation. Turkers unanimously agreed on a single candidate in
87.8\% of the time (across datasets). Finally, as a third quality
filter, we recruited bilingual native speakers to verify, rewrite, and
break ties between all top candidates.  Annotators chose to rewrite
best candidates in only 3.2\% of cases, suggesting our crowdsourced
dataset is well representative of utterances from native
speakers. Example translations from annotators and MT are shown in
Table~\ref{tab:dset_examples}.  Further details of our crowdsourcing
methodology and a sample of human-translated data can be found in
Appendix~\ref{sec:appendix_data}.

\paragraph{Machine Translation}
All machine translation systems used in this work were treated as a
black-box.  For most experiments, we retrieved translations from
English to the target language with the Google Translate API
\citep{gtranslate}. We use this system owing to the purported
translation quality \citep{multilingsp-and-code-switching-Duong2017}
and the API public availability. For ensemble approaches, we used Baidu
Translate and Youdao Translate for Mandarin, and Microsoft Translator
Text and Yandex Translate for German (see Appendix~\ref{sec:appendix_data}).

%%% Local Variables: 
%%% mode: latex
%%% TeX-master: "../../EMNLP2020-Bootstrap/main"
%%% End: 

\begin{table}[t]
\centering
\begingroup
\renewcommand{\arraystretch}{1.3} % Default value: 1

%\resizebox{0.3\textwidth}{!}{%
\begin{small}
\begin{tabular}{@{}rc||c@{}}
\toprule
 & DE & ZH \\ \midrule
\multicolumn{1}{l}{Back-translation to EN} & 53.9 & 57.8 \\
+BERT-base & 56.4 & 58.9 \\ \midrule
\multicolumn{1}{l}{{\sc Seq2Seq}} & 66.9 & 66.2 \\
\multicolumn{1}{r}{+BERT (de/zh)} & 67.8 & 67.4 \\ \midrule
\multicolumn{1}{l}{Shared Encoder} & 69.3 & 68.3 \\
+BERT-ML & \textbf{69.5} & \textbf{68.9} \\ \bottomrule
\multicolumn{3}{c}{(a)~training on gold-standard data}\\
\multicolumn{3}{c}{} \\
\end{tabular}%

\begin{tabular}{@{}rc||c@{}}
\toprule
\multicolumn{1}{l}{} & DE (MT) & ZH (MT) \\ \midrule
\multicolumn{1}{l}{Back-translation to EN} & 57.8 & 53.9 \\
+BERT-base & 58.9 & 56.4 \\ \midrule
\multicolumn{1}{l}{{\sc Seq2Seq}} & 61.0 & 55.2 \\
+BERT-(de/zh) & 64.8 & 57.3 \\ \midrule
\multicolumn{1}{l}{Shared Encoder} & 64.1 & 58.7 \\
+BERT-ML & 66.4 & 59.9 \\ \midrule
\multicolumn{1}{l}{MT-Paraphrase} & 62.2 & 64.5 \\
+BERT-ML & 67.8 & 65.0 \\
+Shared Encoder & 66.6 & 68.1 \\ \midrule
\multicolumn{1}{l}{MT-Ensemble} & 63.9 & 62.2 \\
+BERT-ML & 64.8 & 65.5 \\
+Shared Encoder & \textbf{68.5} & \textbf{68.3} \\ \bottomrule
\multicolumn{3}{c}{(b) training on machine translated (MT) data}
\end{tabular}%
\end{small}
%}
\endgroup
\caption{Test set denotation Accuracy for ATIS in German (DE) and Chinese (ZH).\label{tab:atis_mt_results}}
\end{table}

%%%%%% FIGURE 5: Overnight RESULTS FOR DE(MT) AND ZH(MT)
\begin{table*}[t]
\centering
\begingroup
\renewcommand{\arraystretch}{1.3} % Default value: 1
\resizebox{\textwidth}{!}{%
\begin{tabular}{@{}lccccccccc||ccccccccc@{}}
\toprule
 & \multicolumn{9}{l||}{DE (MT)} & \multicolumn{9}{l}{ZH (MT)} \\ 
 & Ba. & Bl. & Ca. & Ho. & Pu. & Rec. & Res. & So. & Avg. & Ba. & Bl. & Ca. & Ho. & Pu. & Rec. & Res. & So. & Avg. \\ \midrule
Back-translation to EN & 17.6 & 44.1 & 11.3 & 37.0 & 20.5 & 23.1 & 27.4 & 34.0 & 26.9 & 18.2 & 33.6 & 7.7 & 30.2 & 24.2 & 26.9 & 22.3 & 29.4 & 24.1 \\
\multicolumn{1}{r}{+BERT-base} & 59.1 & 51.6 & 28.6 & 38.6 & 29.8 & 37.0 & 32.2 & 60.0 & 42.1 & 47.1 & 33.6 & 33.9 & 34.4 & 33.5 & 36.6 & 27.4 & 52.9 & 37.4 \\ \midrule
{\sc Seq2Seq} & 76.5 & 47.4 & 70.8 & 51.3 & 67.1 & 70.4 & 62.3 & 73.1 & 64.9 & 78.5 & 51.6 & 55.4 & 64.0 & 62.7 & 69.0 & 66.6 & 73.1 & 65.1 \\
\multicolumn{1}{r}{+BERT-(de/zh)} & 74.2 & 56.6 & \textbf{80.4} & 60.8 & 65.8 & 73.6 & 70.8 & 79.2 & 70.2 & \textbf{84.7} & 48.6 & 64.9 & 73.0 & 68.9 & 68.5 & 70.5 & 78.3 & 69.7 \\ \midrule
Shared Encoder & 72.9 & 58.6 & 75.0 & 60.8 & \textbf{76.4} & 73.1 & 63.6 & 75.9 & 69.5 & 78.0 & 46.1 & 61.3 & 67.7 & 65.2 & 70.4 & 63.6 & 76.5 & 66.1 \\
\multicolumn{1}{r}{+BERT-(de/zh)} & 80.8 & 60.4 & 78.6 & 61.4 & 71.4 & \textbf{78.2} & 66.9 & \textbf{79.8} & 72.2 & 81.1 & 51.4 & 66.7 & 71.4 & 65.2 & 67.6 & \textbf{74.7} & 77.5 & 69.4 \\ \midrule 
MT-Paraphrase & 79.5 & 53.4 & 73.8 & 58.7 & 69.6 & 73.1 & 66.9 & 72.4 & 68.4 & 76.0 & 48.6 & 59.5 & 66.7 & \textbf{69.6} & 63.9 & 66.9 & 76.5 & 65.9 \\ 
\multicolumn{1}{r}{+BERT-ML} & 82.4 & 55.4 & 73.8 & \textbf{67.2} & 69.6 & 75.9 & 79.2 & 76.7 & 72.5 & 82.4 & 50.4 & 63.7 & 74.6 & 67.7 & 69.9 & 70.5 & 77.4 & 69.6 \\
\multicolumn{1}{r}{+Shared Encoder} & \textbf{82.6} & 60.7 & 78.6 & 66.1 & 72.0 & 77.3 & 75.0 & 79.2 & 73.9 & 81.3 & 50.9 & \textbf{69.6} & \textbf{75.7} & 65.8 & 72.2 & 69.0 & 77.9 & 70.3 \\ \midrule
MT-Ensemble & 72.1 & 55.8 & 74.1 & 54.4 & 67.9 & 70.2 & 64.9 & 68.6 & 66.0 & 71.1 & 45.8 & 58.3 & 62.2 & 61.5 & 62.0 & 61.1 & 71.4 & 61.7 \\
\multicolumn{1}{r}{+BERT-ML} & 81.0 & 57.3 & 73.9 & 62.2 & 68.3 & 74.2 & \textbf{81.1} & 77.6 & 72.0 & 83.6 & 50.2 & 64.3 & 72.1 & 62.1 & 67.1 & 71.4 & 78.0 & 68.6 \\
\multicolumn{1}{r}{+Shared Encoder} & 81.1 & \textbf{66.7} & 77.9 & 65.9 & 74.4 & 73.1 & 80.4 & 77.5 & \textbf{74.6} & 84.1 & \textbf{52.9} & 69.0 & 74.1 & 65.4 & \textbf{73.6} & 71.1 & \textbf{78.3} & \textbf{71.1} \\ \bottomrule
\end{tabular}
}
\endgroup
\caption{Test set denotation accuracy for Overnight in German (DE) and
  Chinese (ZH) from training on machine translated (MT) data. Results are
  shown for individual domains and an eight-domain average (best
  results in bold). Domains are \textit{Basketball}, \textit{Blocks}, \textit{Calendar}, \textit{Housing}, \textit{Publications}, \textit{Recipes}, \textit{Restaurants} and \textit{Social Network}.\label{tab:onight_mt_results}}
\end{table*}

\section{Results and Analysis}
\label{sec:results}

\begin{figure*}[t!]
    \centering
    \includegraphics[width=\textwidth]{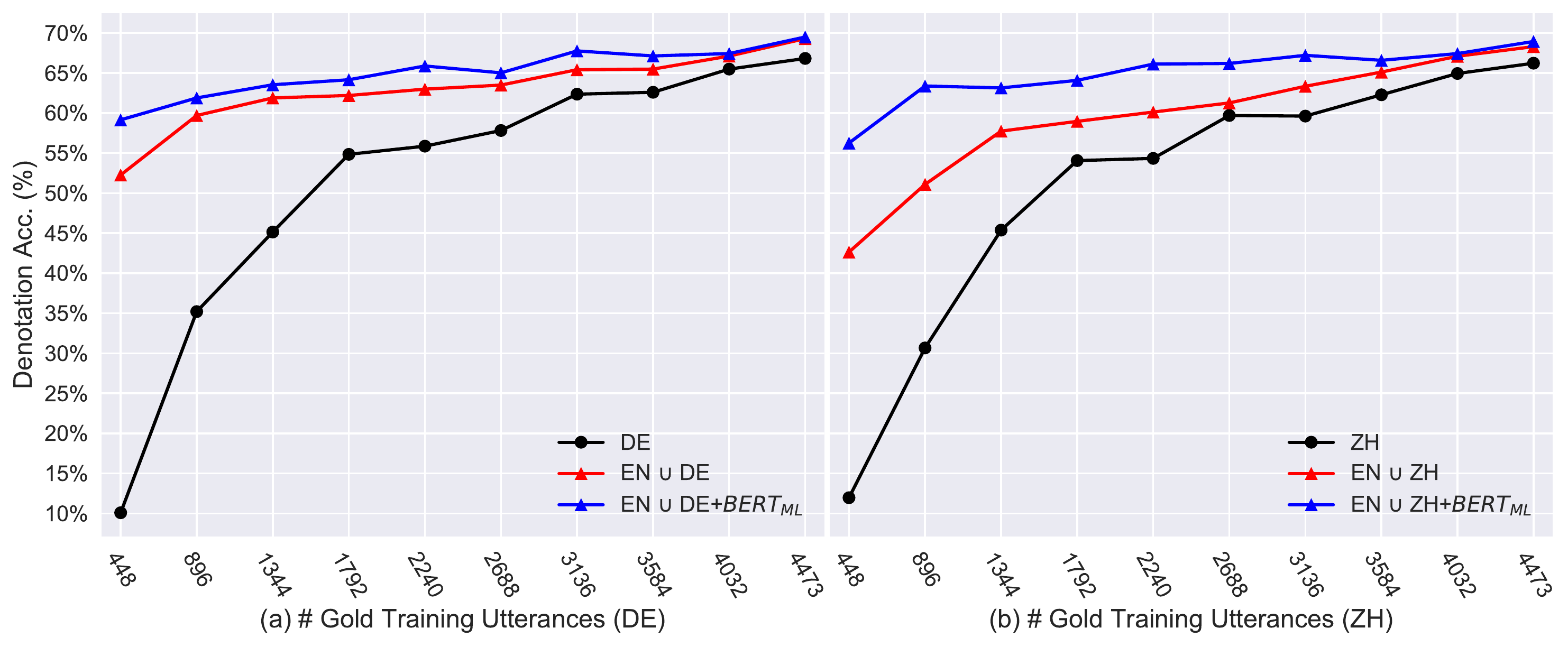} 
\caption{Denotation Accuracy against number of training examples in
  (a) German and (b) Chinese. Augmenting the training data with
  English, $EN\cup{\rm L}$, uses all 4,473 English training utterances
  ($y$ axis shared between figures). Each point averages results on
  three random splits of the dataset.}
    \label{fig:denot_acc_against_n_gold_examples}
\end{figure*}

We compare the neural model defined in Section~\ref{ssec:neural_sp}
({\sc Seq2Seq}) to models using each augmentation outlined in
Section~\ref{sec:feat_aug}, a combination thereof, and the back-translation baseline. Table~\ref{tab:atis_mt_results}(a) details experiments for ATIS using human translated training data, contrasting to Table~\ref{tab:atis_mt_results}(b) which substitutes MT for training data in ZH and DE. Similar results for Overnight are then presented in Table \ref{tab:onight_mt_results}. Finally we consider partial translation in Figure \ref{fig:denot_acc_against_n_gold_examples}. Optimization, hyperparameter settings and reproducibility details are given in Appendix \ref{sec:experimental-setup}.
To the best of our knowledge, we present the first results for executable
semantic parsing of ATIS and Overnight in any language other than
English. While prior multilingual work using $\lambda-$calculus logic
is not comparable, we compare to similar results for English in
Appendix~\ref{sec:appendix:en_results}.

\paragraph{ATIS} 

Table~\ref{tab:atis_mt_results}(a) represents the ideal case of human
translating the full dataset. While this would be the least economical
option, all models demonstrate performance above back-translation with
the best improvement of +13.1\% and +10.0\% for DE and ZH
respectively. This suggests that an in-language parser is preferable
over MT into English given available translations. Similar to
\citet{Shaw2019GeneratingLF-gnn-semparse-acl19} and
\citet{multilingsp-and-code-switching-Duong2017}, we find that
pre-trained BERT representations and a shared encoder are respectively
beneficial augmentations, with the best system using both for ZH and
DE.  However, the latter augmentation appears less beneficial for ZH
than DE, potentially owing to decreased lexical overlap between EN and
ZH (20.1\%) compared to EN and DE (51.9\%). This could explain the
decreased utility of the shared embedding space. The accuracy of our
English model is 75.4\% (see Appendix~\ref{sec:appendix:en_results}),
incurring an upper-bound penalty of~-6.1\% for DE and~-6.5\% for
ZH. Difficulty in parsing German, previously noted by
\citet{multilingual-sp-hierch-tree-Jie2014}, may be an artefact of
comparatively complex morphology. We identified issues similar to
\citet{Min2019-CSPIDER} in parsing Chinese, namely word segmentation
and dropped pronouns, which likely explain weaker parsing compared to
English.

Contrasting to back-translation, the {\sc Seq2Seq} model without BERT
in Table \ref{tab:atis_mt_results}(b), improves upon the baseline by
+3.2\% for DE and +1.3\% for ZH. The translation approach
for German supersedes back-translation for all models, fulfilling the
minimum requirement as a useful parser. However for Chinese, the {\sc
  Seq2Seq} approach requires further augmentation to perform above the
56.4\% baseline. For ATIS, the \mbox{MT-Ensemble}
model, with a shared encoder and BERT-based inputs, yields the best accuracy.
We find that the \mbox{MT-Paraphrase} model performs similarly as a base model
and with pre-trained inputs. As the former model has 3$\times$
the encoder parameters, it may be that additional
data, $\mathcal{D}^{\rm EN}_{\rm train}$, improves each
encoder sufficiently for the \mbox{MT-Ensemble} to improve over smaller models. Comparing between gold-standard human translations, we find similar best-case penalties of -1.0\% for DE and -0.6\% for ZH using MT as training data. The model trained on MT achieves nearly the same generalization error as the model trained on the gold standard. Therefore, we consider the feasibility of our approach justified by this result.

\paragraph{Overnight} 

We now extend our experiments to the multi-domain Overnight dataset,
wherein we have only utterances from native speakers for evaluation,
in Table \ref{tab:onight_mt_results}. Whereas back-translation was
competitive for ATIS, here we find a significant collapse in accuracy
for this baseline. This is largely due to translation errors stemming
from ambiguity and idiomatic phrasing in each locale, leading to
unnatural English phrasing and dropped details in each query. Whereas
\citet{artetxe2020translation-artifacts} found back-translation to be
competitive across 15 languages for NLI, this is not the case for
semantic parsing where factual consistency and fluency in parsed
utterances must be maintained. 

The {\sc Seq2Seq} model with BERT outperforms the baseline by a
considerable +28.1\% for DE and +32.3\% for ZH, further supporting the
notion that an in-language parser is a more suitable strategy for the
task. Our reference English parser attains an average 79.8\% accuracy,
incurring a penalty from crosslingual transfer of -14.9\% for DE and
-14.7\% for ZH with the {\sc Seq2Seq} model. Similar to ATIS, we find
\mbox{MT-Ensemble} as the most performant system, improving over the
baseline by +32.5\% and +33.7\% for DE and ZH respectively. The best
model minimises the crosslingual penalty to -5.2\% for DE and -8.7\%
for ZH. Across both datasets, we find that single augmentations
broadly have marginal gain and combining approaches maximizes
accuracy.

\paragraph{Challenges in Crosslingual Parsing}

We find several systematic errors across our results. Firstly, there are
orthographic inconsistencies between translations that incur sub-optimal learned embeddings. For example, ``5'' can be expressed as
``\begin{CJK*}{UTF8}{gbsn}五\end{CJK*}'' or ``five''. This
issue also arises for Chinese measure words which are often
mistranslated by MT. Multilingual BERT inputs appear
to mostly mitigate this error, likely owing to pre-trained
representations for each fragmented token.

Secondly, we find that multilingual training improved entity
translation errors e.g.~ resolving translations of ``the Cavs'' or
``coach'', which are ambiguous terms for ``Cleveland Cavaliers'' and
``Economy Class''. We find that pairing the training logical form with
the source English utterance allows a system to better disambiguate and
correctly translate rare entities from DE/ZH. This disparity arises during
inference because human translators are more likely to preserve named entities
but this is often missed by MT with insufficient context.

Finally, paraphrasing techniques benefit parsing expressions in DE/ZH
equivalent to peculiar, or KB-specific, English phrases. For example,
the \textit{Restaurants} domain heavily discusses ``dollar-sign'' ratings for
price and ``star sign'' ratings for quality. There is high variation
in how native speakers translate such phrases and subsequently, the
linguistic diversity provided through paraphrasing benefits parsing of
these widely variable utterances.

\paragraph{Partial Translation} 

Our earlier experiments explored the utility of MT for training data,
which assumes the availability of adequate MT. To examine the converse
case, without adequate MT, we report performance with partial
human-translation in
Figure~\ref{fig:denot_acc_against_n_gold_examples}.  Parsing accuracy
on ATIS broadly increases with additional training examples for both
languages, with accuracy converging to the best case performance
outlined in Table~\ref{tab:atis_mt_results}(a).  When translating 50\%
of the dataset, the {\sc Seq2Seq} model performs -10.9\% for DE and
-13.1\% for ZH below the ideal case. However, by using both the shared
encoder augmentation and multilingual BERT ($EN\cup{\rm
  L}+{BERT_{ML}}$), this penalty is minimized to -1.5\% and -0.7\% for
DE and ZH, respectively. While this is below the best system using MT
in Table~\ref{tab:atis_mt_results}(b), it underlines the potential of
crosslingual parsing without MT as future work.

\section{Conclusions}
\label{sec:conclusions}

We presented an investigation into bootstrapping a crosslingual
semantic parser for Chinese and German using only public resources.
Our contributions include a Transformer with attention ensembling and
new versions of ATIS and Overnight in Chinese and German. Our
experimental results showed that a)~multiple MT systems can be queried
to generate paraphrases and combining these with pre-trained
representations and joint training with English data can yield
competitive parsing accuracy; b)~multiple encoders trained with
shuffled inputs can outperform a single encoder; c)~back-translation
can underperform by losing required details in an utterance; and
finally~d) partial translation can yield accuracies $<2\%$ below
complete translation using only 50\% of training data.  Our results
from paraphrasing and partial translation suggest that exploring
semi-supervised and zero-shot parsing techniques is an interesting
avenue for future work.

%%% Local Variables: 
%%% mode: latex
%%% TeX-master: "../../EMNLP2020-Bootstrap/main"
%%% End: 

\paragraph{Acknowledgements}
The authors gratefully acknowledge the support of the UK Engineering and 
Physical Sciences Research Council (grant EP/L016427/1; Sherborne) and 
the European Research Council (award number 681760; Lapata).

% \clearpage
\bibliography{thesis}
\bibliographystyle{acl_natbib}

\section{Appendices}

%\clearpage

\label{sec:appendix}
\appendix
\begin{table*}[!ht]
\centering

\begin{tabular}{@{}l@{~}|@{~}c@{~}c@{~}c@{~}||@{~}l@{~}|@{~}c@{~}c@{~}c@{}}
\toprule
DE & MT1 & MT2 & MT3 & ZH & MT1 & MT2 & MT3 \\ \midrule
G & 0.732 & 0.576 & 0.611 & G & 0.517 & 0.538 & 0.525 \\
MT1 & --- & 0.650 & 0.667 & MT1 & --- & 0.660 & 0.645 \\
MT2 & --- & --- & 0.677 & MT2 & --- & --- & 0.738 \\ \bottomrule
\multicolumn{8}{c}{(a)~ATIS}\\
\multicolumn{8}{c}{} \\
\end{tabular}%

\hspace*{.65cm}\begin{tabular}{@{}l@{~}|@{~}c@{~}c@{~}c@{~}||@{~}l@{~}|@{~}c@{~}c@{~}c@{}}
\toprule
DE & MT1 & MT2 & MT3 & ZH & MT1 & MT2 & MT3 \\ \midrule
MT1 & --- & 0.570 & 0.513 & MT1 & --- & 0.614 & 0.604 \\
MT2 & --- & --- & 0.585 & MT2 & --- & --- & 0.653 \\ \bottomrule
\multicolumn{8}{c}{(b)~Overnight}\\
\end{tabular}%

\vspace*{-2ex}
\caption{Corpus BLEU between gold-standard translations (G) and
  machine translations from sources 1--3 for (a) ATIS and (b)
  Overnight. For German (DE): MT1 is Google Translate, MT2 is
  Microsoft Translator Text and MT3 is Yandex. For Chinese (ZH): MT1
  is Google Translate, MT2 is Baidu Translate and MT3 is Youdao
  Translate.\label{tab:dset_bleu}} 
\end{table*}

\begin{table*}[t!]
\centering
\resizebox{\textwidth}{!}{%
\begin{tabular}{@{}lc||ccccccccc@{}}
\toprule
 & ATIS & \multicolumn{2}{l}{Overnight} &  &  &  &  &  &  &  \\ 
 &  & Ba. & Bl. & Ca. & Ho. & Pu. & Rec. & Res. & So. & Avg \\ \midrule
\citet{Overnight-Wang15} & --- & 46.3 & 41.9 & 74.4 & 54.5 & 59.0 & 70.8 & 75.9 & 48.2 & 58.8 \\
\citet{su-yan-2017-xdomain-paraphrasing} & --- & 88.2 & 62.7 & 82.7 & 78.8 & 80.7 & 86.1 & 83.7 & 83.1 & 80.8 \\
\citet{sp-over-many-kbs-Herzig2017} & --- & 86.2 & 62.7 & 82.1 & 78.3 & 80.7 & 82.9 & 82.2 & 81.7 & 79.6 \\
\citet{iyer-etal-2017-learning-user-feedback} & 82.5 & --- & --- & ---
& --- & --- & --- & --- & --- & --- \\
\citet{wang18-exec-guided-decoding} & 77.9 & --- & --- & --- & --- & --- & --- & --- & --- & --- \\
\citet{iyer-etal-2019-learning-idioms} & {\bf 83.2} & --- & --- & ---
& --- & --- & --- & --- & --- & ---\\
\citet{cao-etal-2019-semantic} & --- & 87.5 & 63.7 & 79.8 & 73.0 & {\bf81.4} & 81.5 & 81.6 & 83.0 & 78.9 \\
\citet{Inan2019ImprovingSP-iclr-reject} & --- & {\bf89.0} & {\bf65.7} & {\bf85.1} & {\bf83.6} & {\bf81.4} & {\bf88.0} & {\bf91.0} & {\bf86.0} & {\bf83.7} \\
\citet{cao-etal-2020-supervised-dual} & --- & 87.2 & {\bf65.7} & 80.4 & 75.7 & 80.1 & 86.1 & 82.8 & 82.7 & 80.1 \\ 
\midrule
{\sc Seq2Seq} & 74.9	&	85.2 &	64.9	&	77.4	&	77.2	&	78.9	&	84.3	&	85.5	&	81.2	&	79.3	\\
\multicolumn{1}{r}{+BERT-base} & 75.4	&	87.7 &	65.4	&	81.0	&	79.4	&	71.4	&	85.6	&	85.8	&	82.0	&	79.8	\\ \bottomrule
\end{tabular}%
}
\caption{Test denotation accuracy on ATIS and Overnight for reference
  model for English. Best accuracy is bolded. Note that  \protect
  \citet{Inan2019ImprovingSP-iclr-reject} evaluate on ATIS, but use
  the non-executable $\lambda-$calculus logical form and are therefore
  not comparable to our results. Domains are \emph{Basketball},
  \emph{Blocks}, \emph{Calendar}, \emph{Housing}, \emph{Publications},
  \emph{Recipes}, \emph{Restaurants},  and \emph{Social Network}.} 
\label{tab:en_results}
\end{table*}

\section{Experimental Setup}
\label{sec:experimental-setup}

For ATIS, we implement models trained on both real and machine-translated 
utterances in German and Chinese. The former is our upper bound, representing
the ideal case, and the latter is the minimal scenario for our developer. 
Comparison between these cases demonstrates both the capability of a system
in the new locale and delineates the adequacy of MT for the task.  
Following this, we explore the multi-domain case of the Overnight dataset
wherein there is no gold-standard training data in either language.

\paragraph{Preprocessing}

Data are pre-processed by removing punctuation and lowercasing with
NLTK \cite{NLTK}, except for cased pre-trained vocabularies and
Chinese. Logical forms are split on whitespace and natural language is
tokenized using the {\tt sentencepiece} tokeniser\footnote{{\tt
    github.com/google/sentencepiece}} to model language-agnostic
subwords. We found this critical for Chinese, which lacks whitespace
delimitation in sentences, and for German, to model word compounding.
For ATIS, we experimented with the entity anonymization scheme from
\citet{iyer-etal-2017-learning-user-feedback}, however, this was found
to be detrimental when combined with pre-trained input representations
and was subsequently not used. 

\paragraph{Evaluation and Model Selection}

Neural models are optimized through a grid search between an embedding/hidden
layer size of $2^{\{7,\ldots10\}}$, the number of layers between \{2,\ldots8\},
the number of heads between \{4,\ldots8\} and the shuffling probability for the \mbox{MT-Ensemble} model between
$p_{\rm shuffle} = \{0.1, \ldots0.5\}$. The best hyperparameters had 6 layers for encoder and decoder, an embedding/hidden layer size of 128, 8 attention heads per layer, a dropout rate of 0.1 and for \mbox{MT-Ensemble} models, we show results for the gated combination approach, which was superior in all cases, and the optimal shuffling probability was 0.4. Models range in size from 4.2-5.7 million parameters. All weights are initialized with Xavier initialization \cite{Glorot10understandingthe} except pre-trained representations which remain frozen. Model weights, $\theta$, are optimized using sequence cross-entropy loss against gold-standard logical forms as supervision. 

Each experiment trains a network for 200 epochs using the Adam Optimizer \citep{ADAMOPT-Kingma2014AdamAM} with a learning rate of 0.001.  We follow the Noam learning rate scheduling approach with a warmup of 10 epochs. Minimum
validation loss is used as an early stopping metric for model selection, with a patience of 30 epochs. 
We use teacher forcing for prediction during training and beam search, with a beam size of 5, during inference. 

Predicted logical forms are input to the knowledge base for ATIS, an
SQL database, and Overnight, SEMPRE \citep{berant-etal-2013-semantic},
to retrieve denotations. All results are reported as exact-match (hard)
denotation accuracy, the proportion of predicted logical forms which 
execute to retrieve the same denotation as the reference query. 
Models are built using PyTorch \citep{pytorch}, AllenNLP \citep{AllenNLP} and HuggingFace BERT models \citep{Wolf2019HuggingFacesTS}. Each parser is trained using a cluster of 16 NVIDIA P100 GPUs with 16GB memory, with each model demanding 6-16 hours to train on a single GPU.

\section{English Results}
\label{sec:appendix:en_results}

We compare our reference model for English to prior work in
Table~\ref{tab:en_results}.
Our best system for this language uses the {\sc Seq2Seq} model outlined in Section~\ref{ssec:neural_sp} with input features from the pre-trained BERT-base model. We acknowledge
our system performs below the state of the art for ATIS by -7.8\% and
Overnight by -3.9\%, but this is most likely because we omit any
English-specific feature augmentation other than BERT. In comparison
to prior work, we do not use entity anonymization, paraphrasing,
execution-guided decoding or a mechanism to incorporate feedback for
incorrect predictions from humans or neural critics. The closest
comparable model to ours is reported by
\citet{wang18-exec-guided-decoding}, implementing a similar {\sc
  Seq2Seq} model demonstrating 77.0\% test set accuracy. However, this
result uses entity anonymization for ATIS to replace each entity with a
generic label for the respective entity type. Prior study broadly found
this technique to yield improved parsing accuracy \citep{iyer-etal-2017-learning-user-feedback,
  seq2tree-Dong2016-lang2logic,
  finegan-dollak-etal-2018-improving-text-to-sql}, a crosslingual
implementation requires crafting multiple language-specific
translation tables for entity recognition. We attempted to implement
such an approach but found it to be unreliable and largely
incompatible with the vocabularies of pre-trained models.

\section{Data Collection}
\label{sec:appendix_data}

\paragraph{Translation through Crowdsourcing}
For the task of crosslingual semantic parsing, we consider the ATIS
dataset \citep{atis-Dahl:1994:ESA:1075812.1075823} and the Overnight
dataset \citep{Overnight-Wang15}.  The former is a single-domain
dataset of utterances paired with SQL queries pertaining to a database
of travel information in the USA. Overnight covers eight domains using
logical forms in the $\lambda-$DCS formalism
\citep{dcs-Liang:2013:LDC:2483810.2483815} which can be executed in
the SEMPRE framework \citep{berant-etal-2013-semantic}.

ATIS has been previously translated into Chinese and Indonesian for the study
of semantic parsing into $\lambda-$calculus logical forms \citep{arch-for-neural-multisp-Susanto2017}, 
however Overnight exists only in English. To the best of our knowledge, there is presently no multi-domain dataset for executable semantic parsing in more than two languages. 
As previously mentioned in Section \ref{sec:data}
, we consider Chinese and German 
in this paper to contrast between a language similar and dissimilar to English and also due to the reported availability of crowd-sourced workers for translation \citep{langdemographicsofturk-Pavlick2014}
and bilingual native speakers for verification. 

To facilitate task evaluation in all languages of interest, we
require a full parallel translation of ATIS in German, for comparison
to the existing Chinese implementation, and a partial translation of
Overnight in both German and Chinese. 
For task evaluation in all languages, we require a full parallel translation of ATIS to complement the existing Chinese translation from \citep{arch-for-neural-multisp-Susanto2017}. As previously discussed, we translate only the development and test set of Overnight \citep{Overnight-Wang15} into Chinese and German for assessment of crosslingual semantic parsing in a multi-domain setting.  Therefore, we translate all 5,473 utterances in ATIS and 4,311 utterances in Overnight. 
The original Overnight dataset did not correct spelling errors from
collected English paraphrases, however, we consider it unreasonable to
ask participants in our task to translate misspelled words, as
ambiguity in correction could lead to inaccurate translations. We
subsequently identified and corrected spelling errors using word
processing software.

We use Amazon Mechanical Turk (MTurk) to solicit three
translations per English source sentence from crowdsourced workers
(Turkers), under the assumption that this will collect at least one
adequate translation \citep{fastcheapcreative-Callison-Burch2009}.
Our task design largely followed practices for translation without
expert labels on MTurk
\citep{crowdsourcingfromnonprofessionals-Zaidan2011,
  parallel6indianlangs-Post2012,
  improvingeduMTwithcrowdsourcing-Behnke2018,multilingualeducorpus-Sosoni2018}.
The task solicits translations by asking a Turker to translate 10
sentences and answer demographic questions concerning country of origin 
and native language. Submissions were restricted to Turkers from Germany,
Austria and Switzerland or China, Singapore, and the USA for German and Chinese respectively. 
We built an AMT interface with quality controls which restricted Turkers
from inputting whitespace and disabled copy/paste anywhere
within the webpage. Attempting to copy or paste in the submission
window triggered a warning that using online translation tools will
result in rejection. Inauthentic translations were rejected if they
held an $>$80\% average BLEU to reference translations from Google
Translate \citep{gtranslate}, as were nonsensical or irrelevant
submissions. For the Chinese data collection, we also rejected
submissions using Traditional Chinese Characters or Pinyin
romanization. Instructions for the initial candidate collection task
are given in Figure~\ref{fig:turk_instr_task1} and the ranking task
in Figure~\ref{fig:turk_instr_task2}. We found 94\% of workers completed the optional demographic survey and that all workers reported their first language Chinese or German as desired. For Chinese, 94\% of workers came from the USA and reported to have spoken Chinese for $>$20 years, and remaining workers resided in China. For German, all workers came from Germany and had spoken German for $>$25 years.

Turkers submitted 10 translations per task for~\$0.7 and \$0.25 to
rank 10 candidate translations, at an average rate to receive an
equivalent full-time wage of \$8.23/hour. This is markedly above the
average wage for US workers of \$3.01/hour discovered by
\citet{Hara2019-worker-pay-demog-turk-chi}. To ensure data quality and
filter disfluencies or personal biases from Turkers, we then recruited
bilingual postgraduate students, native speakers of the task language,
to judge if the best chosen translation from Turk was satisfactory or
required rewriting. If an annotator was dissatisfied with the
translation ranked best from Turk then they provided their own, which
only occurred for 3.2\% of all translations.  Verifiers preferred the
MT candidate over the Turk submissions for 29.5\% of German rankings
and 22.6\% of Chinese rankings, however, this preference bias arose 
only in translations of small sentences (five or fewer words) where
MT and the Turk translation were practically identical. We paid~\$12 an hour for this verification but to minimize cost, we did not collect multiple judgments per translation. We found that verification was completed at a rate of 60 judgments per hour, leading to an approximate cost of \$2200 per language for Overnight and \$2500 for ATIS into German. While this may be considered expensive, this is the minimum cost to permit comparable evaluation in every language. Sample translations for ATIS into German are given in Table \ref{tab:atis_de_sample} and sample translations
for Overnight into German and Chinese are given in Table \ref{tab:onight_zhde_sample}.

\paragraph{Machine Translation}
In this work, we evaluate the feasibility of using machine translation (MT)
as a proxy to generate in-language training data for semantic parsing
of two languages. All MT systems are treated as black-box models without
inspection of underlying translation mechanics or recourse for correction. For most experiments in this work, we use translations from English to the target language using Google Translate \citep{gtranslate}. We use this system owing to the purported translation quality \citep{multilingsp-and-code-switching-Duong2017} 
and because the API is publicly available, contrasting to the closed MT used in \citet{conneau2018xnli}. 

Additionally, we explore two approaches to modeling an ensemble of
translations from multiple MT sources. We expect, but cannot
guarantee, that each MT system will translate each utterance
differently for greater diversity in the training corpus overall. 
For this approach, we consider two additional MT
systems each for Chinese and German. For Mandarin, we use Baidu
Translate and Youdao Translate. For German, we use Microsoft
Translator Text and Yandex Translate.  To verify that the ensemble of
multiple MT systems provides some additional diversity, we
measure the corpus level BLEU between training utterances from each
source. These scores for ATIS, with comparison to human translation, and Overnight are detailed in Table
\ref{tab:dset_bleu}.

Overall, we find that each MT system provides a different set of
translations, with no two translation sets more similar than
any other. We also find that for ATIS in German,
\citet{gtranslate} provides the most similar training dataset to the
gold training data. However, we find that Microsoft Translator Text
appears to narrowly improve translation into Chinese by $+$0.021
BLEU. This arises as an effect of a systematic preference for a polite
form of Chinese question, beginning with
\begin{CJK*}{UTF8}{gbsn}``请''\end{CJK*} [please], preferred by the
professional translator. Overall, we collected all training data using MT for $<\$50$ across both datasets and languages. 

\begin{table*}[ht]
\centering
\resizebox{\textwidth}{!}{%
\begin{tabular}{@{}l@{}}
\toprule
\textbf{Translate all 10 sentences into Simplified Chinese}  \\ \midrule
In this task, we ask you to provide a translation into Simplified Chinese of an English question. \\
You must be {\bf native speaker of Chinese (Mandarin) and proficient in English} to complete this HIT.\\ 
We ask you to use {\bf only Simplified Chinese characters} (\begin{CJK*}{UTF8}{gbsn}简体汉字\end{CJK*}) and {\bf do not use Pinyin} (\begin{CJK*}{UTF8}{gbsn}汉语拼音\end{CJK*}).\\ Attempt to translate every word into Chinese. If this is difficult for rare words you do not understand,\\ such as a person's name or place names, then please copy the English word into the translation. \\ 
You can assume all currency amounts are US Dollars and all measurements are in feet and inches.\\
In order to receive payment, you must complete all translations without using online translation services. \\
The use of online translation websites or software will be considered cheating. \\
Identified cheating will result in withheld payment and a ban on completing further HITs. \\
The demographic questionnaire is optional and you are welcome to complete as many HITs as you like. \\ \bottomrule
\end{tabular}
}
\captionof{figure}{Instructions provided to Turkers for the English to Chinese translation task of Overnight \citep{Overnight-Wang15}. We specify the requirement to answer in Simplified Chinese characters and specify the basis for rejection of submitted work. Instructions are condensed for brevity.} \label{fig:turk_instr_task1}
\end{table*}

\begin{table*}
\centering
\resizebox{\textwidth}{!}{
\begin{tabular}{@{}l@{}}
\toprule
\textbf{Select the best German translation for 10 English sentences}  \\ \midrule

In this HIT, you will be presented with an English question and three candidate translations \\
of this English sentence in German. We ask you to use your judgment as a native-speaker of  \\
German to select the best German translation from the three candidates. \\
If you consider all candidate translations to be inadequate, then provide your own translation. \\
You must be {\bf native speaker of German and proficient in English} to complete this HIT.\\ 
We consider the best translation as one which asks the same question in the style of a native \\
speaker of German, rather than the best direct translation of English. Occasionally, multiple \\
candidates will be very similar, or identical, in this case select the first identical candidate. \\
You must complete all 10 to submit the HIT and receive payment.\\
You are welcome to submit as many HITs as you like. \\
\bottomrule
\end{tabular}
}
\captionof{figure}{Instructions provided to Turkers for the English to German translation ranking for both ATIS \citep{atis-Dahl:1994:ESA:1075812.1075823} and Overnight\citep{Overnight-Wang15}. Instructions are condensed for brevity.} \label{fig:turk_instr_task2}
\end{table*}

\begin{table*}[ht]
\resizebox{\textwidth}{!}{
\begin{tabular}{@{}p{10cm}p{10cm}@{}}
\toprule
English & Translation into German \\ \midrule
What ground transportation is available from the Pittsburgh airport to the town? & Welche Verkehrs Anbindung gibt es vom Pittsburgh Flughafen in die Stadt? \\
Could you please find me a nonstop flight from Atlanta to Baltimore on a
Boeing 757 arriving at 7pm? & Könntest du für mich bitte einen Direktflug von Atlanta nach Baltimore auf einer Boeing 757 um 19 Uhr ankommend finden? \\
What is fare code QO mean? & Was bedeutet der ticketpreiscode QO? \\
Show me the cities served by Canadian Airlines International. & Zeige mir die Städte, die von den Canadian Airlines International angeflogen werden. \\
Is there a flight tomorrow morning from Columbus to Nashville? & Gibt es einen Flug morgen fr{\"u}h von Columbus nach Nashville? \\
Is there a Continental flight leaving from Las Vegas to New York nonstop? & Gibt es einen Continental-flug ohne Zwischenstopps, der von Las Vegas nach New York fliegt? \\
I would like flight information from Phoenix to Denver. & Ich hätte gerne Informationen zu Fl{\"u}gen von Phoenix nach Denver. \\
List flights from Indianapolis to Memphis with fares on Monday. & Liste Fl{\"u}ge von Indianapolis nach Memphis am Montag inklusive ticketpreisen auf. \\
How about a flight from Milwaukee to St. Louis that leaves Monday night? & Wie w{\"a}re es mit einem Flug von Milwaukee nach St. Louis, der Montag Nacht abfliegt? \\
A flight from St. Louis to Burbank that leaves Tuesday afternoon. & Einen Flug von St. Louis nach Burbank, der Dienstag Nachmittag abfliegt. \\ \bottomrule
\end{tabular}
}
\caption{Sample translations from English to German for the ATIS dataset \citep{atis-Dahl:1994:ESA:1075812.1075823}.\label{tab:atis_de_sample}}
\end{table*}

\begin{table*}[ht]
\resizebox{\textwidth}{!}{
\begin{tabular}{@{}p{10cm}p{10cm}@{}}
\toprule
English & Translation into German \\ \midrule
What kind of cuisine is Thai Cafe? & Welche Art von Küche bietet das Thai Café? \\
What neighborhood has the largest number of restaurants? & Welche Wohngegend hat die meisten Restaurants? \\
Which recipe requires the longest cooking time? & Welches Rezept benötigt die längste Kochzeit? \\
Which player had a higher number of assists in a season than Kobe Bryant? & Welcher Spieler hatte eine höhere Anzahl an Vorlagen in einer Saison als Kobe Bryant? \\
Housing with monthly rent of 1500 dollars that was posted on January 2? & Welche Wohnung hat eine monatliche Miete von 1500 Dollar und wurde am 2. Januar veröffentlicht? \\
What article is cited at least twice? & Welcher Artikel wurde mindestens zweimal zitiert? \\
What block is to the right of the pyramid shaped block? & Welcher Block befindet sich rechts neben dem pyramidenförmigen Block? \\
What is the birthplace of students who graduated before 2002? & Was ist der Geburtsort von Studenten, die vor 2002 ihren Abschluss gemacht haben? \\
Who is the shortest person in my network? & Wer ist die kleinste Person in meinem Netzwerk? \\
Find me the employee who quit between 2004 and 2010. & Welche Angestellten haben zwischen 2004 und 2010 gekündigt? \\ \midrule
English & Translation into Chinese \\ \midrule
Hotels that have a higher rating than 3 stars? & { \begin{CJK*}{UTF8}{gbsn}评级高于3星级的酒店\end{CJK*}} \\
Thai restaurants that accept credit cards? & { \begin{CJK*}{UTF8}{gbsn}接受信用卡的泰式餐馆\end{CJK*}} \\
Show me recipes posted in 2004 or in 2010? & { \begin{CJK*}{UTF8}{gbsn}告诉我2004年或2010年发布的食谱\end{CJK*}} \\
Which player has played in fewer games than Kobe Bryant? & { \begin{CJK*}{UTF8}{gbsn}哪个球员比科比布莱恩特打得比赛少？\end{CJK*}} \\
Meeting that has duration of less than three hours? & { \begin{CJK*}{UTF8}{gbsn}时长短于3小时的会议\end{CJK*}} \\
Meetings in Greenberg Cafe that end at 10am? & { \begin{CJK*}{UTF8}{gbsn}在Greenberg咖啡厅举行并且在早上10点结束的会议\end{CJK*}} \\
Housing units that are smaller than 123 Sesame Street? & { \begin{CJK*}{UTF8}{gbsn}比123芝麻街要小的房屋单元\end{CJK*}} \\
Publisher of article citing Multivariate Data Analysis? & { \begin{CJK*}{UTF8}{gbsn}引用多变量数据分析的文章出版商\end{CJK*}} \\
Block that is below at least two blocks? & { \begin{CJK*}{UTF8}{gbsn}在至少两个块以下的块\end{CJK*}} \\
Find me all students who attended either Brown University or UCLA. & { \begin{CJK*}{UTF8}{gbsn}给我找到所有要么在布朗大学要么在UCLA上学的学生们\end{CJK*}}
\end{tabular}

}
\caption{Sample translations from English to German and Chinese for the Overnight dataset \citep{Overnight-Wang15}.\label{tab:onight_zhde_sample}}
\end{table*}

%%% Local Variables: 
%%% mode: latex
%%% TeX-master: "../acl2020"
%%% End: 

\end{document}